%% file: main.tex
\documentclass[10pt,twocolumn,letterpaper]{article}

\usepackage{cvpr}              %
\usepackage{graphicx}
\usepackage{amsmath}
\usepackage{amssymb}
\usepackage{booktabs}
\usepackage{soul}
\usepackage{paralist}

\usepackage[pagebackref,breaklinks,colorlinks]{hyperref}

\usepackage[capitalize]{cleveref}
\crefname{section}{Sec.}{Secs.}
\Crefname{section}{Section}{Sections}
\Crefname{table}{Table}{Tables}
\crefname{table}{Tab.}{Tabs.}

\input{notations}
\input{shortcuts}

\begin{document}

\title{\methodname: Towards Generative Detailed Neural Avatars}

\author{\vspace{.05cm} Xu Chen$^{1,3}$ \quad Tianjian Jiang$^{1}$ \quad Jie Song$^{1}$ \quad Jinlong Yang$^{3}$\\ \vspace{.2cm} Michael J. Black$^{3}$ \quad Andreas Geiger$^{2,3}$ \quad Otmar Hilliges$^{1}$\\
\vspace{.05cm}$^1$ETH Z{\"u}rich, Department of Computer Science \quad $^2$University of Tübingen \\ $^3$Max Planck Institute for Intelligent Systems, T{\"u}bingen\\
\vspace{.1cm}\href{https://xuchen-ethz.github.io/gdna}{https://xuchen-ethz.github.io/gdna}
}

\twocolumn[{%
\renewcommand\twocolumn[1][]{#1}%
\vspace{-5em}
\maketitle
\vspace{1em}
\input{figures/teaser}
}
]
\begin{abstract}
\input{sections/0_abstract}

\end{abstract}
\input{sections/1_introduction}
\input{sections/2_related_work}

\input{sections/3_method}

\input{sections/4_experiments}

\input{sections/5_conclusion}
\input{sections/6_acknowledgement}

{\small
\bibliographystyle{ieee_fullname}
\bibliography{bibliography_long,bibliography,bibliography_custom}
}

\end{document}

%% file: notations.tex
\newcommand{\point}{\mathbf{x}}
\newcommand{\latent}{\mathbf{z}}

\newcommand{\mfunction}[1]{\mathbf{#1}}

\newcommand{\raydir}{\mathbf{r}_d}
\newcommand{\rayori}{\mathbf{r}_c}

\newcommand{\pixel}{\mathbf{p}}
\newcommand{\loss}{\mathcal{L}}
\newcommand{\bodypose}{\boldsymbol{\theta}}
\newcommand{\bodyshape}{\boldsymbol{\beta}}

\newcommand{\bone}{\boldsymbol{B}}

\newcommand{\occ}{o}

\newcommand{\norm}{\mathbf{n}}
\newcommand{\skin}{\mathbf{w}}

\newcommand{\focc}{\mathcal{O}}

\newcommand{\fnorm}{\mathcal{N}}
\newcommand{\fskin}{\mathcal{W}}
\newcommand{\fwarp}{\mathcal{M}}

\newcommand{\zocc}{\latent_\text{shape}}

\newcommand{\znorm}{\latent_\text{detail}}

\newcommand{\feat}{\mathbf{f}}

\newcommand{\R}[1]{\mathbb{R}^{#1}}

%% file: shortcuts.tex
\newcommand{\figref}[1]{Fig.~\ref{#1}}
\newcommand{\secref}[1]{Section~\ref{#1}}

\newcommand{\eqnref}[1]{Eq.~\eqref{#1}}
\newcommand{\tabref}[1]{Tab.~\ref{#1}}

\makeatletter
\DeclareRobustCommand\onedot{\futurelet\@let@token\@onedot}
\def\@onedot{\ifx\@let@token.\else.\null\fi\xspace}
\def\eg{e.g\onedot} 
\def\ie{i.e\onedot} 
\def\cf{cf\onedot}

\def\etal{et~al\onedot}

\makeatother

\newcommand{\boldparagraph}[1]{\vspace{0.2cm}\noindent{\bf #1:}}

\usepackage{xcolor}
\definecolor{turquoise}{cmyk}{0.65,0,0.1,0.3}
\definecolor{purple}{rgb}{0.65,0,0.65}
\definecolor{dark_green}{rgb}{0, 0.5, 0}
\definecolor{orange}{rgb}{0.8, 0.6, 0.2}
\definecolor{red}{rgb}{0.8, 0.2, 0.2}
\definecolor{darkred}{rgb}{0.6, 0.1, 0.05}
\definecolor{blueish}{rgb}{0.0, 0.3, .6}
\definecolor{light_gray}{rgb}{0.7, 0.7, .7}
\definecolor{pink}{rgb}{1, 0, 1}
\definecolor{greyblue}{rgb}{0.25, 0.25, 1}

\newif\ifshowcomments
\showcommentstrue %

\ifshowcomments

    \newcommand{\oh}[1]{ \noindent {\color{red} {\bf OH:} {#1}} }
    \newcommand{\ag}[1]{ \noindent {\color{red} {\bf AG:} {#1}} }
    \newcommand{\mjb}[1]{ \noindent {\color{red} {\bf MJB:} {#1}} }
    \newcommand{\jy}[1]{ \noindent {\color{red} {\bf JY:} {#1}} }
    \newcommand{\xc}[1]{ \noindent {\color{red} {\bf XC:} {#1}} }
 \else
    \newcommand{\oh}[1]{\unskip}
    \newcommand{\ag}[1]{\unskip}
    \newcommand{\mjb}[1]{\unskip}
    \newcommand{\jy}[1]{\unskip}
    \newcommand{\xc}[1]{\unskip}
\fi   

\newcommand{\update}[1]{{\color{black}{#1}}}

\newcommand{\methodname}{gDNA\xspace}
\newcommand{\suppmat}{Sup.~Mat.\xspace}

%% file: figures/teaser.tex
\vspace{-4em}
\begin{center}
    \includegraphics[width=\textwidth,trim=40 20 50 10, clip]{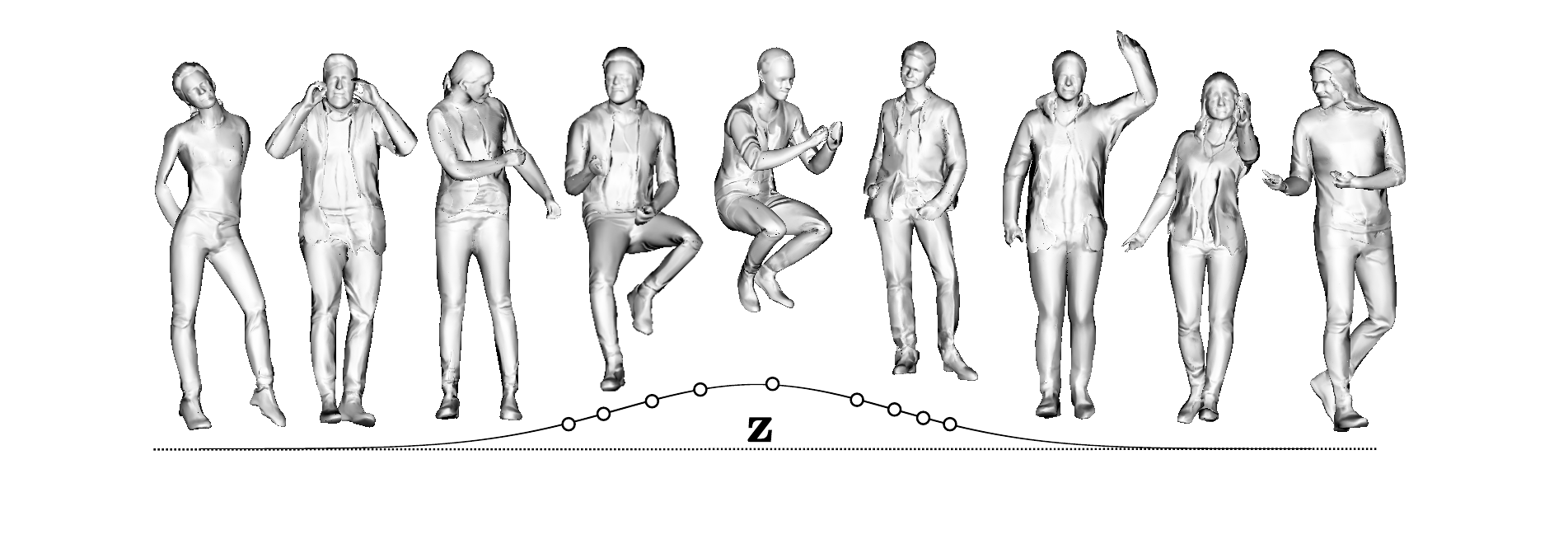}
    \vspace{-3.2em}
    \captionof{figure}{\textbf{Generative Detailed Neural Avatars.} We propose a method to \emph{generate} 1) a diverse set of 3D virtual humans of 2) varied identity, gender and shape, appearing in 
    3) different clothing styles and poses, with
    4) realistic and stochastic details such as wrinkles in garments. Our multi-subject method learns shape, articulation and clothing details from few posed scans without requiring skinning weight supervision. The method is able to synthesize novel identities that are not in the training set and generalizes to unseen poses.}
    \label{fig:teaser}
\end{center}

%% file: sections/0_abstract.tex
To make 3D human avatars widely available, we must be able to generate a variety of 3D virtual humans with varied identities and shapes in arbitrary poses.
This task is challenging due to the diversity of clothed body shapes, their complex articulations, and the resulting rich, yet stochastic geometric detail in clothing. 
Hence, current methods that represent 3D people do not provide a full generative model of people in clothing.
In this paper, we propose a novel method that learns to generate detailed 3D shapes of people in a variety of garments with corresponding skinning weights.
Specifically, we devise a multi-subject forward skinning module that is learned from only a few posed, un-rigged scans per subject.
To capture the stochastic nature of high-frequency details in garments, we leverage an adversarial loss formulation that encourages the model to capture the underlying statistics. 
We provide empirical evidence that this leads to realistic generation of local details such as wrinkles. 
We show that our model is able to generate natural human avatars wearing diverse and detailed clothing. 
Furthermore, we show that our method can be used on the task of fitting human models to raw scans, outperforming the previous state-of-the-art.

%% file: sections/1_introduction.tex
\section{Introduction}

The ability to easily create diverse high-quality virtual humans with full control over their pose has many applications in movie production, games, VR/AR, architecture, and computer vision. While modern computer graphics techniques achieve photorealism, they typically require a lot of expertise and extensive manual effort. Our goal is to make 3D human avatars widely accessible by learning a {\em generative model of people}. Towards this goal, we propose the first method that can \emph{generate} 
\begin{inparaenum}[1)]
    \item diverse 3D virtual humans with 
    \item various identities and shapes, appearing in 
    \item different clothing styles and poses, with
    \item realistic and stochastic high-frequency details such as wrinkles in garments. 
\end{inparaenum}

Generative modeling of 3D rigid objects has recently seen rapid progress, fueled by continuous and resolution-independent neural 3D representations \cite{Mescheder2019CVPR, Park2019CVPRDeepsdf,chen2019CVPRimnet,Schwarz2020NEURIPS, niemeyer2021giraffe}.
However, modeling clothed humans and their articulation is more difficult due to the complex interaction of garments, their topology, and pose-driven deformations. 
Recent work leverages neural implicit surfaces to learn high-quality articulated avatars for a \emph{single} subject \cite{Deng2020ECCV,Saito2021CVPR,Chen2021ICCV,Tiwari2021ICCVNeuralGIF} but these methods are not generative, \ie they cannot synthesize novel human identities and shapes.
Generative models of \emph{clothing} exist that augment SMPL by predicting displacements from the body mesh (CAPE~\cite{Ma2020CVPR}), or by draping an implicit garment representation on a T-posed body (SMPLicit~\cite{Corona2021CVPRSMPLicit}), and relying on SMPL's learned skinning for reposing. We show empirically that holistic modeling of identity, shape, articulation and clothing leads to higher fidelity generation and animation of virtual humans and to higher accuracy in fitting to 3D scans.

Taking a step towards fully \textbf{g}enerative modeling of \textbf{d}etailed \textbf{n}eural \textbf{a}vatars, we propose \methodname, a method that synthesizes 3D surfaces of \emph{novel} human shapes, with control over the clothing style and pose, and that produces realistic high-frequency details of the garments (Fig.~\ref{fig:teaser}). 
To leverage raw (\emph{posed}) 3D scans, we build a multi-subject implicit generative representation. 
We build upon SNARF \cite{Chen2021ICCV}, a recent method for learning \emph{single-subject} articulation-dependent effects that has been shown to generalize well to unseen poses. 
SNARF~\cite{Chen2021ICCV} requires \emph{many} poses of a \emph{single} subject for training. 
In contrast, our \emph{multi-subject} method can be learned from \emph{very few} posed scans (1-3) of many different subjects. 
This is achieved via the addition of a latent space for the conditional generation of shape and skinning weights for clothed humans. 
Furthermore, a learned warping field yields accurate deformations, using the same skinning field, independent of body size. 

Clothing wrinkles are produced by an underlying stochastic process. To capture these effects, we propose a method that learns the underlying statistics of 3D clothing details via an adversarial loss. 
Previous mesh-based approaches formulate this in UV-space \cite{laehner2018deepwrinkles}, which is not directly applicable to implicit surfaces due to the lack of mesh connectivity.
\update{To learn high-frequency details, we first predict a 3D normal field, conditioned on the coarse shape features. 
To backpropagate the adversarial loss to the 3D normal field we establish 3D-2D correspondences by augmenting forward skinning with an implicit surface renderer.
We show that adversarial training leads to significantly improved fidelity of 3D geometric details, see Fig.~\ref{fig:result_ablation}.
}

Trained from posed scans only, we demonstrate the first method that can generate a large variety of 3D clothed human shapes with detailed wrinkles under pose control. The generated samples can be reposed via the learned skinning weights. 
We evaluate \methodname quantitatively, qualitatively, and through a perceptual  study; \methodname strongly outperforms baselines. 
Furthermore, we show that \methodname can be used for fitting and re-animation of 3D scans,\update{ outperforming the state of the art (SOTA). In summary, we contribute:
\begin{itemize}
\setlength\itemsep{-.3em}
    \item The first method to generate a large variety of animatable 3D human shapes in detailed garments; that
    \item learns from raw posed 3D scans without requiring canonical shapes, detailed surface registration, or manually defined skinning weights, and
    \item a technique to significantly improve the geometric detail in clothing deformation, based on recovering the underlying statistics of cloth deformation.
\end{itemize}
}

%% file: sections/2_related_work.tex
\section{Related Work}

\input{figures/pipeline}

\boldparagraph{2D and 3D Generative Models}
Most modern methods for synthesizing natural images leverage generative adversarial networks (GANs)~\cite{Goodfellow2014NIPS} or variational auto-encoders (VAEs)~\cite{kingma2013auto}. These methods have achieved a high level of photorealism~\cite{karras2019style,karras2020analyzing,Karras2021} and can yield impressive results on the task of synthesizing 2D images of humans~\cite{Lassner:GeneratingPeople:2017,NIPS2017_6644,chen2019unpaired, Grigorev2021CVPRStylePeople,sarkar2021humangan,albahar2021pose}.
However, such methods reason in 2D and hence 3D consistency cannot be guaranteed ~\cite{Lassner:GeneratingPeople:2017,NIPS2017_6644,chen2019unpaired} nor is extracting 3D geometry from such approaches straightforward.

Several methods for the task of learning \emph{rigid} 3D shapes exist. 
Early methods rely on voxel~\cite{wu2016NIPS3dgan} or point cloud~\cite{achlioptas2018ICMLlearning} representations. 
More recently, several methods represent object shapes by learning an implicit function using neural networks~\cite{Park2019CVPRDeepsdf, Mescheder2019CVPR, chen2019CVPRimnet}. 
Such representations have also been proposed for the task of \emph{generative} modeling of 3D shapes \cite{Mescheder2019CVPR, Park2019CVPRDeepsdf,chen2019CVPRimnet,nguyen2019hologan,chen2020category,Schwarz2020NEURIPS,niemeyer2021giraffe,devries2021unconstrained}. However, these methods are typically not easily extended to non-rigid clothed humans. In this paper, we study the problem of 3D implicit \emph{generative} modeling of \emph{non-rigid} human shape.

\boldparagraph{3D Human Models}
Parametric 3D human body models~\cite{Loper2015SIGGRAPH, Anguelov2005SIGGRAPH,joo2018total, osman2020star, Xu2020CVPR} can synthesize 3D human shapes from a set of low-dimensional control parameters by deforming a template mesh. This idea has also been extended to model clothed humans~\cite{alldieck2018video,Ma2020CVPR}. However, geometric expressivity is limited due to the fixed mesh topology and the bounded resolution of the template mesh. 

To overcome the topology and resolution limitations of meshes, other representations, including point clouds~\cite{Ma2021CVPR, Ma2021ICCV,zakharkin2021point}, implicit surfaces~\cite{Saito2021CVPR, Chen2021ICCV,Tiwari2021ICCVNeuralGIF,shao2021doublefield, wang2021metaavatar,Palafox2021ICCV}, and radiance fields \cite{peng2021animatable,2021narf,su2021nerf,xu2021h,liu2021neural}, have been explored. 
In particular, neural implicit surface representations have emerged as a powerful tool to model 3D (clothed) human shapes~\cite{Saito2019ICCV,Saito2020CVPR,Deng2020ECCV,he2020geopifu,li2020monocular,zheng2021pamir,Mihajlovic2021CVPR,he2021arch++,alldieck2021imghum,xiu2022icon,dong2022pina,zheng2022avatar} due to their topological flexibility and resolution independence.
Recent work~\cite{Saito2021CVPR, Chen2021ICCV, Tiwari2021ICCVNeuralGIF} uses implicit surfaces to learn human avatars for a \emph{single} subject, wearing a \emph{specific} garment. These methods model clothing details such as wrinkles as a deterministic function of the body poses. However, due to hysteresis and complex material properties, garment folds and wrinkles are stochastic and existing methods struggle to capture these effects. %
In contrast, we propose a \emph{multi-subject} generative model of 3D humans that provides separate control over poses, garments and can synthesize realistic geometric details.

CAPE~\cite{Ma2020CVPR} and SMPLicit~\cite{Corona2021CVPRSMPLicit} are generative models of \emph{clothing} only, based on meshes and implicit surfaces respectively. Both methods are purely additive, that is they drape an implicit garment over the SMPL body~\cite{Corona2021CVPRSMPLicit} or predict the displacement parameters of a SMPL+D template mesh~\cite{Ma2020CVPR}. We experimentally show that this leads to lower fidelity in generated samples and higher error when fitting to 3D scans.   
NPMs~\cite{Palafox2021ICCV} provide a latent space of multiple subjects for fitting to RGB-D depth maps or 3D scans.

A common problem of all aforementioned approaches is the specific training data requirements these models impose: 
They either require synthetic data in canonical space~\cite{Corona2021CVPRSMPLicit,Palafox2021ICCV}, or precise registration of a template mesh to posed scans~\cite{Ma2020CVPR,Palafox2021ICCV}. The former are rare and suffer from a domain gap, while the latter is challenging to attain.
Our method overcomes this issue by requiring only a \textit{few} training samples of each subject in \textit{posed} space. We show that our method learns complex shape and clothing details and models realistic deformation even from such limited data.

\boldparagraph{Adversarial Training of Clothing Details}
Adversarial loss formulations have been used to learn detailed cloth wrinkles by optimizing 2D representations such as UV normal maps~\cite{laehner2018deepwrinkles} or depth images~\cite{wang2020ECCVnormalgan}. It is noteworthy that implicit surfaces lack a notion of connectivity and therefore, incorporating 2D representations that have been designed to augment explicitly parametrized meshes is not straightforward. In contrast, we propose a formulation that leverages a 2D adversarial loss computed with posed images to optimize a 3D implicit representation in canonical space. Finally, our focus is the generation of human shapes appearing in varied clothing styles and diverse identities while previous methods focus on reconstruction~\cite{wang2020ECCVnormalgan} or single garment pose-dependent wrinkle enhancement~\cite{laehner2018deepwrinkles}.

%% file: figures/pipeline.tex
\begin{figure*}
    \centering
    \includegraphics[width=\textwidth,trim=30 0 0 0, clip]{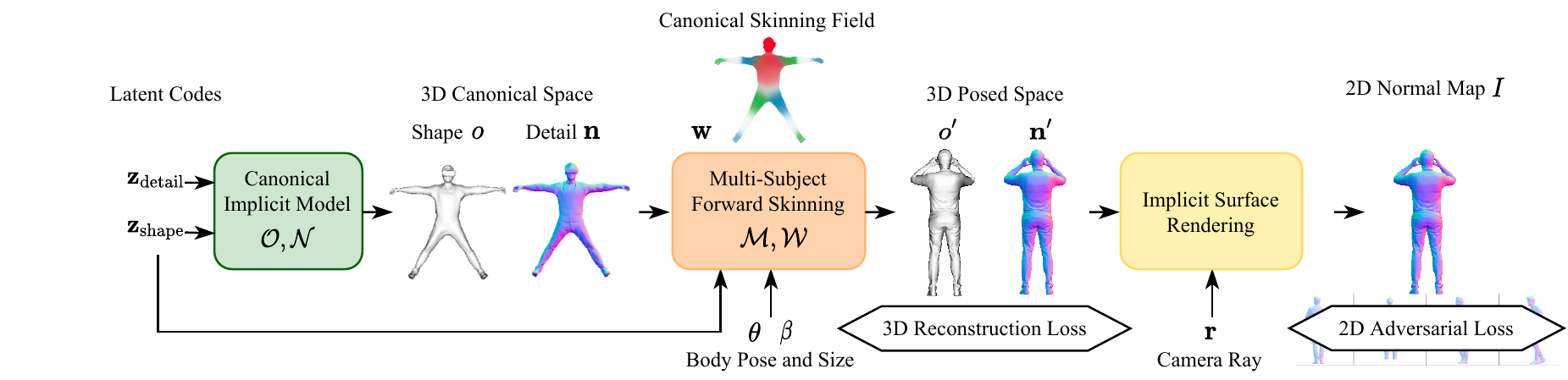}
    \vspace{-2em}
    \caption{\textbf{Method Overview.} We represent clothed humans using coarse shapes and detailed normals in a pose- and body-size-independent canonical space. This canonical representation can then be deformed to target body poses and sizes via a multi-subject forward skinning module. The deformed shapes are compared to raw posed scans via a 3D reconstruction loss to learn canonical shapes and skinning. To improve surface details, we augment the forward skinning module with an implicit surface renderer to generate 2D normal maps and learn detailed 3D normal fields by applying a 2D adversarial loss formulation. }
    \vspace{-1em}
    \label{fig:pipeline}
\end{figure*}

%% file: sections/3_method.tex
\section{Method}
Our goal is to build a model that generates diverse 3D clothed humans with varying identities and fine-grained geometric details in arbitrary poses. Our model is learned from a sparse set of static scans without assuming surface correspondences. Our method is summarized \figref{fig:pipeline}.

First, we formulate a pose- and body-size-independent canonical representation of clothed human shapes (\secref{sec:model}).
Second, to learn the canonical shape and deformation properties from very few posed scans of each of the subjects, we extend a single-subject differentiable forward skinning method~\cite{Chen2021ICCV} to multiple subjects via a latent space of shape, articulation and garment (\secref{sec:deform}). 
Finally, to learn rich yet stochastic geometric details, we learn a detailed 3D normal field via a 2D adversarial loss formulation. 
To achieve this, we augment the forward skinning module with an implicit surface renderer (\secref{sec:render}). Training details are discussed in \secref{sec:train}.

\subsection{Canonical Representation}
\label{sec:model}
Our method is based on neural implicit representations, leveraging their topological flexibility and resolution independence. We model the clothed human shape and geometric clothing details jointly.

\boldparagraph{Coarse Shape} We model the shape in canonical space as the $\tau=0.5$ level set of a neural occupancy function:
\begin{align}
\mathcal{S}(\zocc) = \{ \point \mid \focc(\point,\zocc) = \tau\} ,
\end{align}
where $\focc$ is a neural network that predicts the occupancy probability $\occ$ for any 3D point $\point$ in canonical space. The prediction is conditioned on a \emph{shape} code $\zocc \in  \R{L_\text{shape}} $: 
\begin{align}
\focc: \R{3} \times \R{L_\text{shape}}  &\rightarrow [0,1] \times \R{L_{\feat}} \\
        (\point,\zocc)  &\mapsto (\occ, \feat) \nonumber
\label{eqn:occ}
\end{align}
This occupancy network also outputs a feature vector $\feat$ of dimension $L_{\feat}$ for each surface point. This feature carries coarse shape information and is used to predict fine details. 

\input{figures/method_canonical}

We combine a 3D CNN-based feature generator and a locally conditioned MLP to model $\focc$. A 3D style-based~\cite{nguyen2019hologan,karras2019style} generator, illustrated in \figref{fig:network} first produces a 3D feature volume conditioned on $\zocc$ via adaptive instance normalization~\cite{huang2017arbitrary}. The final occupancy is obtained via trilinear sampling of the feature volume and by feeding the feature and the 3D coordinate into an MLP.

\boldparagraph{Detailed Surface Normals} Learning an occupancy field for multiple subjects and garment types with accurate and detailed normals is challenging and we empirically show that a na\"ive implementation leads to artifacts on the surface (\cf \figref{fig:result_ablation}). Analogous to normal mapping for polygon meshes \cite{blinn1978simulation,laehner2018deepwrinkles}, we model surface details via normals in canonical 3D space. Such surface normals can be represented by the gradient of the implicit function, but this results in considerable computational complexity. Therefore, we use an MLP to predict surface normals similar to \cite{Tiwari2021ICCVNeuralGIF}. However, because implicit surfaces have no notion of connectivity, we propose a geometry-aware approach to link coarse geometry and the detailed normal field. More specifically, we condition the surface normal prediction on the underlying shape, leveraging the feature $\feat$ from the occupancy network. We further condition the field on a latent $\znorm \in \R{L_\text{detail}}$ to enable generation of controllable details for the same coarse shape:
\begin{align}
\fnorm: \R{3} \times \R{L_\text{detail}} \times \R{L_{\feat}} &\rightarrow \R{3} \\
        (\point,\znorm,\feat)  &\mapsto \norm  \nonumber
\end{align}

\subsection{Multi-Subject Forward Skinning}
\label{sec:deform}

We additionally model the deformation properties and define the body size ($\bodyshape$) and pose ($\bodypose$) parameters to be consistent with SMPL, enabling use of existing datasets (\eg AMASS \cite{Mahmood2019ICCV}) for animation. The body size parameter $\bodyshape$ is a 10-dimensional vector, and the body pose parameter $\bodypose$ represents the joint angles of SMPL's skeleton.

\boldparagraph{Single-Subject Skinned Representation} 
To animate implicit human shapes in controllable body poses $\bodypose$, recent work \cite{Mihajlovic2021CVPR,Saito2021CVPR,Chen2021ICCV,Tiwari2021ICCVNeuralGIF} generalizes mesh-based linear blend skinning algorithms to neural implicit surfaces. The skeletal deformation of each 3D point is modeled as the weighted average of a set of bone transformations, with weights at each point predicted by an MLP. A key difference is whether this skinning weight field is defined in canonical space or in posed space. 
We follow Chen \etal~\cite{Chen2021ICCV} who define the skinning field in canonical space: 
\begin{align}
\fskin: \R{3} &\rightarrow \R{n_b} \\
        \point  &\mapsto \skin  \nonumber ,
\end{align}
where $n_b$ denotes the number of bones and the weights $\mathbf{w}=\{w_1,\dots,w_{n_b}\}$ of each point $\point$ are enforced to satisfy $w_i \geq 0$ and $\sum_{i} w_i = 1$ by a softmax activation function. As shown in \cite{Chen2021ICCV}, defining the skinning weights field in canonical space is desirable because the skinning weights are then pose-independent, thus easier to learn and enabling generalization to out-of-distribution poses.

\input{figures/method_skinning}

\boldparagraph{Multi-Subject Skinned Representation} 
We extend this forward skinning idea to multiple subjects.
Since the skinning weight field is defined in canonical space, the model can aggregate information over multiple training instances.
Importantly, this enables us to learn skinning from \emph{one or a few} poses of \emph{multiple} subjects, instead of requiring \emph{many} poses of the \emph{same} subject.  

To achieve this, we decouple the effects originating from the body size variation $\bodyshape$ and the clothed human shape $\zocc$. 
We model the skinning field in a body-size-neutral space, analogously to the canonical surface representations. To capture diverse clothed human shapes, we condition the field on the latent shape code $\zocc$:
\begin{align}
\fskin: \R{3} \times \R{L_\text{shape}} &\rightarrow \R{n_b} \\
        (\point, \zocc)  &\mapsto \skin  \nonumber
\label{eqn:skin}
\end{align}

We then model body size change with an additional warping field. Given a point $\hat\point$ in $\bodyshape$-size space, the warping field maps it back to the mean size by predicting its canonical correspondence $\point$ (see \figref{fig:skinning}): 
\begin{align}
\fwarp: \R{3} \times \R{L_{\bodyshape}} &\rightarrow \R{3} \\
        (\hat\point,\bodyshape)  &\mapsto \point  \nonumber
\label{eqn:warp}
\end{align}
\update{In this formulation, $\bodyshape$ captures body shape variations analogously to SMPL, \eg body height. Therefore, the canonical shape network only needs to model the remaining shape variations beyond SMPL, \eg clothing and hair, controlled by $\zocc$}. The final resized canonical surface is defined by:
\begin{align}
\hat{\mathcal{S}}(\zocc,\bodyshape) = \{ \hat\point \mid \focc(\fwarp(\hat\point, \bodyshape),\zocc) = \tau\}
\end{align}
Given the target body pose $\bodypose$, a point $\hat\point$ in $\bodyshape$-size space is transformed to posed space $\point'$ via
\update{
\begin{align}
\point' &= \mathbf{d}(\point, \bodyshape, \bodypose, \zocc) \nonumber \\ &=\textstyle \sum_{i=1}^{n_{\text{b}}}\fskin_{i}(\fwarp(\hat\point,\bodyshape),\zocc) \cdot \bone_i(\bodyshape, \bodypose)\cdot \hat\point ,
\label{eqn:lbs}
\end{align}}where $\bone_i(\bodyshape, \bodypose)$ are the bone transformation matrices obtained from the parametric skeleton of SMPL.

\boldparagraph{Implicit Differentiable Forward Skinning} 
While our model learns a canonical representation, its supervision is provided in posed space. Given a point $\point'$ in posed space we need to determine its correspondence in canonical space $\point$ to compare the predicted occupancy and normals to ground-truth. We first find the correspondence $\hat\point^*$ of $\point'$ in resized canonical space and then map $\hat\point^*$ to canonical space $\point^*$. An overview is provided in \figref{fig:skinning}. While the goal is to determine $\point' \mapsto \hat\point$, we only have direct access to the inverse mapping defined by forward skinning \eqnref{eqn:lbs}, which is not invertible. Following \cite{Chen2021ICCV}, we determine the correspondence numerically by finding the root of the equation:
\begin{align}
\mathbf{d}(\hat\point, \bodyshape, \bodypose, \zocc) - \point' = \mathbf{0},
\label{eqn:root_equ}
\end{align}
using Broyden's method \cite{Broyden1965BOOK}.
Subsequently, the canonical correspondence $\point^*$ is given by: 
\begin{align}
\point^* &= \fwarp(\hat\point^*,\bodyshape)
\label{eqn:correspondence}
\end{align}
We can now determine the occupancy at $\point'$ as $\occ'= \focc(\point^*, \zocc)$ and the normal $\norm'$  as
\update{
\begin{align}
\norm' = (\textstyle\sum_{i=1}^{n_{\text{b}}}\fskin_{i}(\point^*,\zocc) \cdot \mathbf{R}_i)^{-T} \fnorm(\point^*, \feat, \znorm)
\label{eqn:lbs_norm}
\end{align}
}
where $\mathbf{R}_i$ denotes the rotational component of $\bone_i$.

For convenient future reference, we define the occupancy field $\focc'$ and normal function $\fnorm'$ in posed space as:
\begin{align}
\focc'&: (\point',\zocc,\bodyshape,\bodypose)  \mapsto \occ', \feat \\
\fnorm'&: (\point',\znorm,\feat,\bodyshape,\bodypose)  \mapsto \norm'
\end{align}

\subsection{Implicit Surface Rendering}
\label{sec:render}
Geometric clothing details are challenging to learn due to their stochastic nature. In 2D image generation tasks, GANs have achieved impressive results on learning high fidelity local textures. We propose to learn better geometric details $\fnorm$ using an adversarial loss. Towards this goal, we augment the forward skinning module with an implicit renderer to establish direct correspondences between 2D projections of 3D points in posed space and corresponding 3D points in canonical space, enabling end-to-end training.

\boldparagraph{Implicit Rendering with Skinning} Given a pixel $\pixel$ in the 2D posed normal map, its correspondence in deformed 3D space $\point'$ can be determined by the intersection between the ray through $\pixel$ and the forward skinned surface: 
\begin{align}
\focc'(\point',\zocc,\bodyshape,\bodypose) = \tau, \text{with \ }\point' = \rayori + t \cdot \raydir
\end{align}
where $\raydir$ and $\rayori$ denote the ray direction and origin, and $t$ is the scalar distance along the ray. Following \cite{Niemeyer2020CVPR}, we determine the intersection point $\point'$ by finding the first change of occupancy $\focc'$ along the ray using the Secant method. We also obtain the canonical correspondence point $\point$ of $\pixel$ via forward skinning. Solving the 3D canonical correspondence for each pixel, yields the 2D normal map $I$:  
\begin{align}
I_{\pixel} = \fnorm'(\point',\znorm,\feat,\bodyshape,\bodypose)
\end{align}

\input{figures/result_sample}
\vspace{-1em}
\subsection{Training}
\label{sec:train}
We train our method via a set of posed scans and their corresponding SMPL parameters $\bodypose,\bodyshape$. We follow the auto-decoding framework of \cite{Park2019CVPRDeepsdf}, and assign one shape code $\zocc$ and one detail code $\znorm$ to each training sample. These are initialized to be zero and optimized jointly with the network weights. To enable sampling, we fit a Gaussian distribution to the latent codes after training.

We split training into two stages: We first train the coarse shape, skinning, and warping networks and then train the normal network. This two-stage training is essential. Otherwise, the normal supervision will be back-propagated to wrong locations in canonical space due to wrong correspondences before training of shape and skinning converges.

For the first stage, we use the binary cross entropy loss $\loss_\text{BCE}$ between predicted occupancy $\focc'(\point',\zocc,\bodyshape,\bodypose)$ and ground-truth $\occ_\text{gt}$.
\update{Following \cite{Chen2021ICCV}, we add auxiliary losses $\loss_\text{bone}$ and $\loss_\text{joint}$ to guide learning during early iterations:
\begin{align}
&\loss_\text{bone} = BCE(\focc(\point_\text{bone},\zocc),1) \\
&\loss_\text{joint} = \| \mfunction{w}(\point_\text{joint},\zocc) - \mathbf{w}_\text{joint, target} \|_2^2
\end{align}
where $\point_\text{bone}$ are randomly sampled points on canonical bones, $\point_\text{joint}$ are randomly sampled canonical joints, and $\mathbf{w}_\text{joint, target}$ is a vector that is $0.5$ for the neighboring bones and $0$ elsewhere (for details see \suppmat).}
To ensure that the warping field changes body size consistently, we enforce the warping field to warp SMPL vertices $\mathbf{v}(\bodyshape)$ to the corresponding location in the neutral shape $\mathbf{v}(\bodyshape_0)$:
\begin{align}
\loss_\text{warp} = \| \fwarp( \mathbf{v}(\bodyshape), \bodyshape)  - \mathbf{v}(\bodyshape_0) \|_2^2
\end{align}
Finally, we regularize the latent code to be close to the origin of the latent space via $\loss_\text{reg,shape} = \| \zocc \|_2^2$.

The normal prediction network is trained subsequently. Here we penalize differences between the predicted and GT normal $ \norm'_\text{gt}$ for randomly sampled surface points:
\begin{align}
\loss_\text{norm} = 1 - {\norm'_\text{gt}}^T \cdot \fnorm'(\point',\znorm,\feat,\bodyshape,\bodypose)
\end{align}
\update{In addition, we apply non-saturating adversarial losses \cite{Goodfellow2014NIPS} $\loss_\text{adv} = -\log(1+\exp(D(I)))$ with $R_1$ gradient penalty \cite{Mescheder2018ICML} on the predicted 2D normal maps $I$ and the real normal maps rendered from the posed scans $I_\text{real}$. $D$ is a jointly trained discriminator (see \suppmat for details). We further regularize $\znorm$ with $\loss_\text{reg,detail} = \| \zocc \|_2^2$.}

\subsection{Inference}
\vspace{-0.5em}
\update{We generate human avatars by randomly sampling $\zocc$ and $\znorm$ from the estimated Gaussian distribution. We then extract meshes in resized canonical space using MISE~\cite{Mescheder2019CVPR} from the implicit representation $\hat{\mathcal{S}}(\zocc,\bodyshape)$ and predict the vertex normal with our normal field. Finally, we pose the meshes to desired poses $\bodypose$ following \eqnref{eqn:lbs}. }

%% file: figures/method_canonical.tex
\begin{figure}
    \centering
    \includegraphics[width=0.8\linewidth,trim=40 0 20 0, clip]{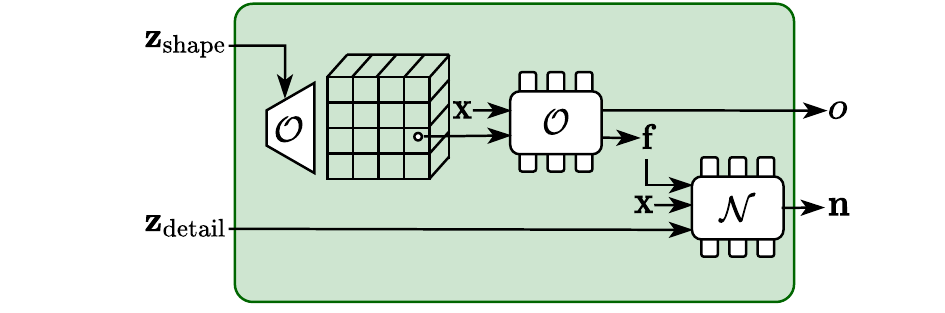}
    \caption{\textbf{Canonical Implicit Model.} Given latent codes $\zocc$ and $\znorm$,  this module predicts the occupancy probability $\occ$ and the normals $\norm$ for 3D points in canonical space. To predict $\occ$, a 3D style-based generator first produces a 3D feature volume conditioned on $\zocc$. The final occupancy of a 3D point is obtained via trilinear sampling of the feature volume and by feeding the feature and the 3D coordinate into an MLP. To predict normals $\norm$, we use an MLP conditioned on feature $\feat$ and latent code $\znorm$.}
    \label{fig:network}
    \vspace{-0.3cm}
\end{figure}

%% file: figures/method_skinning.tex
\begin{figure}
    \centering
    \includegraphics[width=\linewidth,trim=40 0 20 15, clip]{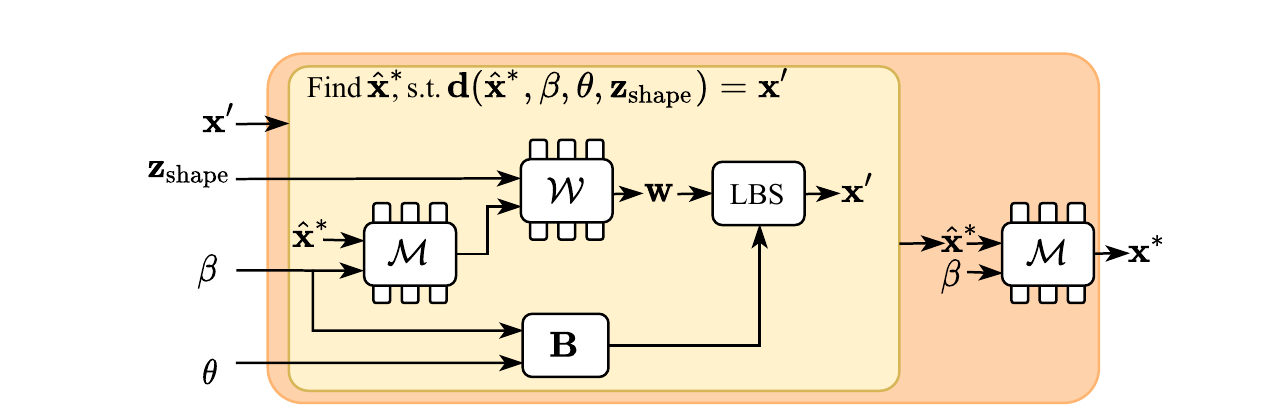}
    \caption{\textbf{Multi-subject Forward Skinning.} This module deforms the canonical occupancy and normal field to the target pose $\bodypose$ and size $\bodyshape$ by establishing correspondences from posed space $\point'$ to canonical space $\point^*$. First, given a deformed point $\point'$, we find its correspondence in resized canonical space $\hat\point^*$ by iteratively finding the root of \eqref{eqn:root_equ}. Subsequently, we map $\hat\point^*$ to $\point^*$ in size-neutral canonical space using the warping field. }
    \label{fig:skinning}
    \vspace{-0.3cm}
\end{figure}

%% file: figures/result_sample.tex
\begin{figure*}
    \centering
    \includegraphics[width=\linewidth,trim=15 30 10 00, clip]{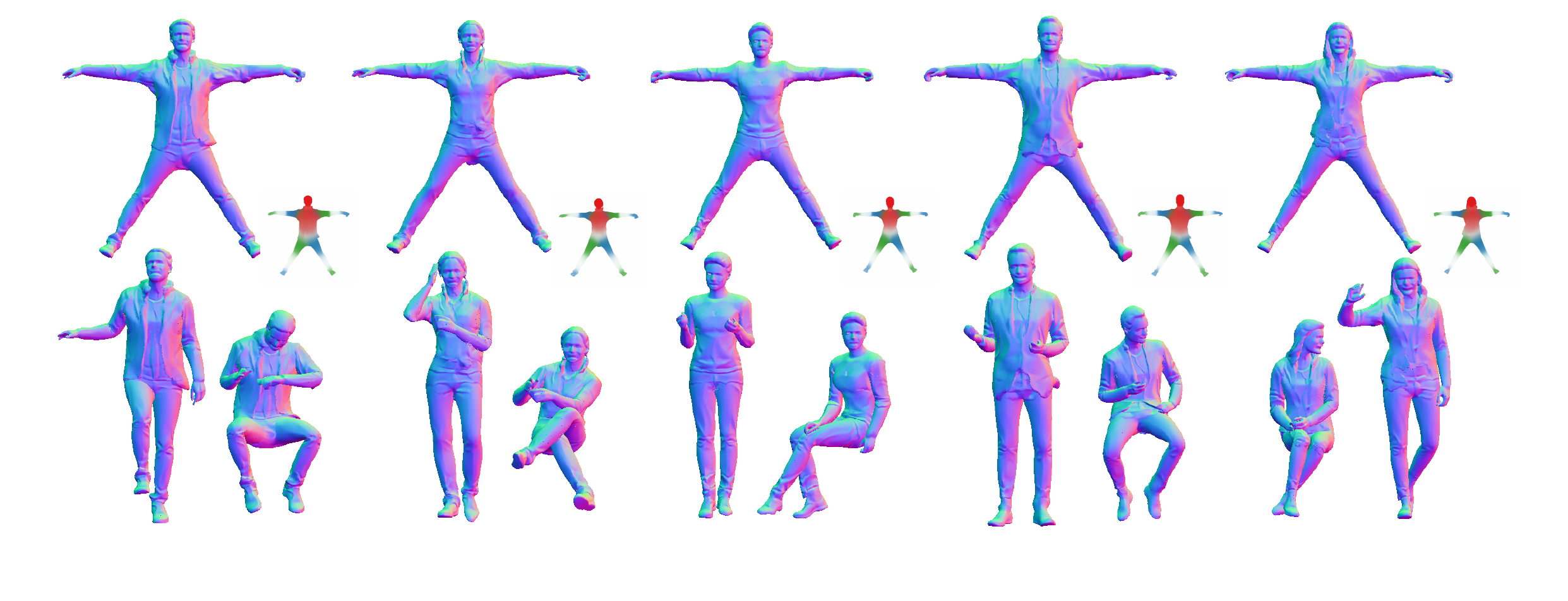}
    \vspace{-1.7em}
    \caption{\textbf{Clothed Human Generation.} We show randomly sampled clothed human shapes generated by our method. 
    \emph{Top:} generated canonical shapes with detailed normals and skinning weights. \emph{Bottom:} generated shapes reposed (2 poses each). Shapes are visualized as normal maps to highlight the detailed geometry synthesized by our method. For shaded results, see other figures, \eg \figref{fig:teaser}.
    }
    \label{fig:result_sample}
    \vspace{-1.2em}
\end{figure*}

%% file: sections/4_experiments.tex
\section{Experiments}
\vspace{-0.5em}
\update{Our main goal is to generate 3D human avatars. Since we are the first to tackle this problem setting, we compare our method to carefully designed ablative baselines, enabling analysis of each component of our method. We also evaluate the expressiveness of our model by fitting it to unseen scans and compare the accuracy to SOTA 3D human shape modeling methods.
We outline the evaluation protocols in the following and refer the readers to \suppmat for details.}

\boldparagraph{\underline{Datasets}}

\boldparagraph{3D Scans} We train our model on commercial scans \cite{WEBRenderPeople,WEB3DPeople}.

\boldparagraph{SIZER} Following \cite{Corona2021CVPRSMPLicit}, we use the SIZER dataset~\cite{Tiwari2020ECCV} to evaluate fitting. This dataset contains 3D scans of humans in 21 garments, including shirts, T-shirts, coats and pants.

\boldparagraph{\underline{Metrics}}

\update{\boldparagraph{Fr\'echet Inception Distance (FID)} To evaluate generation quality, we compute FID between 2D normal maps of training scans and those of randomly generated 3D shapes.}

\update{\boldparagraph{User Preference} We conduct a perceptual study among 44 subjects and report how often participants preferred a particular method over ours.}

\boldparagraph{Surface Distance} To evaluate fitting accuracy, we measure the  one-directional Chamfer distance between predicted surfaces and the target scans, following SMPLicit~\cite{Corona2021CVPRSMPLicit}.

\boldparagraph{\underline{Baselines}}

\update{\boldparagraph{NPMs~\cite{Palafox2021ICCV}} NPMs learn the latent space of human shapes and deformation from ground-truth canonical shapes and vertex displacements, obtained from synthetic 3D animations \cite{li20214dcomplete} and real scans with registered surfaces \cite{Ma2020CVPR}. }

\update{\boldparagraph{SMPLicit~\cite{Corona2021CVPRSMPLicit}} SMPLicit learns a generative model of 3D garments, and drapes these over the SMPL T-pose. This model is trained with a collection of 3D synthetic garments. }

\vspace{-1em}
\subsection{Quality of Generated Samples}
\boldparagraph{Random Generation of Canonical Shapes} We show random samples generated by our method in \figref{fig:result_sample} (top). While trained with posed scans only, our method learns plausible canonical shapes with surface details.

\input{figures/result_interp}

\boldparagraph{Disentangled Pose and Shape} The generated shapes can be reposed as desired, even to poses far beyond the training pose distribution (\cf \figref{fig:result_sample} bottom and \figref{fig:teaser}). 

\boldparagraph{Interpolation} Interpolating the shape and details codes, yields smooth transitions of shapes and details between two very different samples, as shown in \figref{fig:result_interp}.

\boldparagraph{Disentangled Shape and Details} Our disentangled formulation allows us to generate diverse clothing details for the same coarse shape. \figref{fig:result_detail} shows results with the same coarse shape $\zocc$ but different details codes $\znorm$. While the coarse shape remains the same, \methodname generates varied plausible wrinkles that match the underlying coarse shape.

\input{figures/result_detail}
\input{figures/result_nn}

\boldparagraph{Extrapolation Beyond Training Distribution} To further illustrate generalization, we show the training samples with the most similar pose and latent code to the generated sample in \figref{fig:result_nn}. The nearest neighbors are noticeably different from our generation, demonstrating that our method generalizes and is able to generate novel shapes in novel poses.

\subsection{Ablation Study}
\vspace{-0.5em}
We now ablate our design choices. The results are summarized in \tabref{tab:ablations} and \figref{fig:result_ablation}.
\vspace{-0.2em}

\boldparagraph{Canonical Space Modeling}
We verify the necessity to model shapes in canonical space and joint learning of skinning weights. Towards this goal, we implement a baseline that generates posed shapes directly given the latent code and the body pose as input. As shown in \figref{fig:result_ablation} (first row), the individual samples lack details as the baseline must capture a large shape space caused by the pose change. Since the method does not reason about articulation, the sampled shapes suffer from invalid pose configurations, leading to high FID values as shown in \tabref{tab:ablations} (\emph{Pose ONet}).
\vspace{-0.2em}

\boldparagraph{Adversarial Learning}
The adversarial loss plays an important role in improving the perceptual realism of the generated samples, as evidenced by the FID improvement from \emph{Detailed Normal (w/o Adversarial)} to \emph{Ours} in \tabref{tab:ablations}. The normals estimated directly from the occupancy field suffer from artifacts on the surface (\figref{fig:result_ablation} (second row)). Training without adversary leads to overly smooth geometry (\figref{fig:result_ablation} (third row)), as the reconstruction loss induces a bias that
averages out details. In contrast, our method produces realistic high-frequency details (\figref{fig:result_ablation} (bottom)). Notably, in 21.3\% of the cases, users consider our generated shapes to be even more realistic than real scans.

\subsection{Comparison with SOTA on Model Fitting}
\vspace{-0.5em}
While our main goal is to generate clothed human shapes, our model can be fit to raw observations, just like existing 3D parametric human or clothing models. We consider two recent SOTA methods, \ie NPMs~\cite{Palafox2021ICCV} and SMPLicit~\cite{Corona2021CVPRSMPLicit}. We follow SMPLicit~\cite{Corona2021CVPRSMPLicit} and fit ours and the baselines to scans from the SIZER dataset.

\boldparagraph{Accuracy} While not designed for fitting, our method achieves better accuracy than previous special purpose methods, as demonstrated in \tabref{tab:fitting}. \update{Our method captures the person identity and clothing shapes more faithfully than NPMs and SMPLicit, and our results exhibit more details such as wrinkles (\figref{fig:result_repose_fitting} top). 
Since the model is trained directly from posed scans, disentangling pose and shape, it learns about real clothing details and can reproduce them.
}

\update{\boldparagraph{Reposing Scans} During fitting we also recover skinning weights. This enables  reposing of the shape as demonstrated in \figref{fig:result_repose_fitting} bottom.}
\input{tables/ablation}
\input{figures/result_ablation}

%% file: figures/result_interp.tex
\begin{figure*}
    \centering
    \includegraphics[width=\linewidth,trim=20 0 10 20, clip]{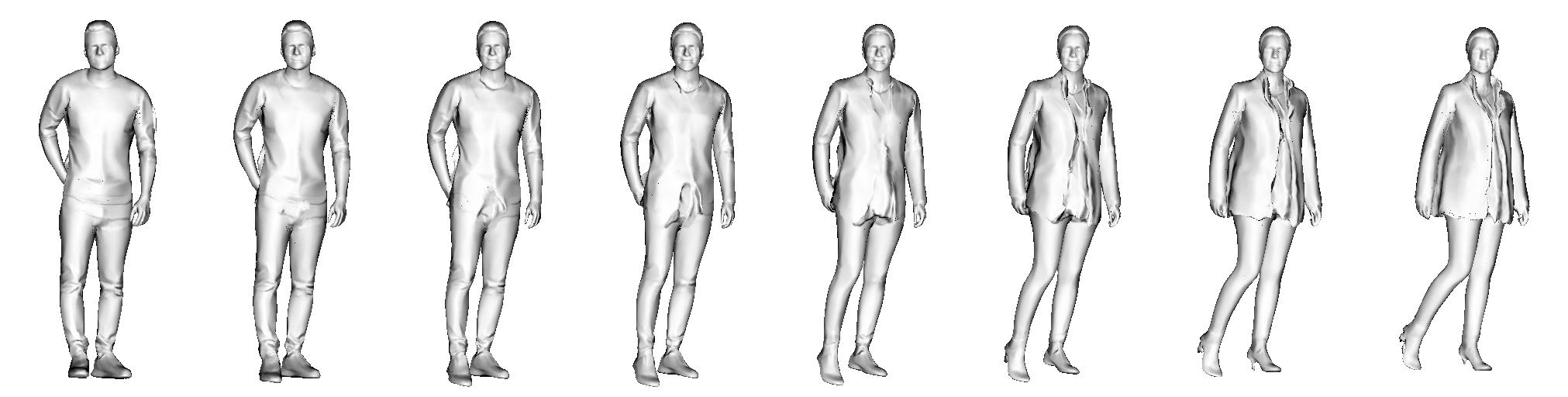}
    \vspace{-2.5em}
    \caption{\textbf{Interpolation.} We interpolate the pose and shape code and the detail code between the leftmost and rightmost sample. }
    \label{fig:result_interp}
    \vspace{-1em}
\end{figure*}

%% file: figures/result_detail.tex
\begin{figure}[t!]
    \centering
    \includegraphics[width=\linewidth,trim=20 10 10 0, clip]{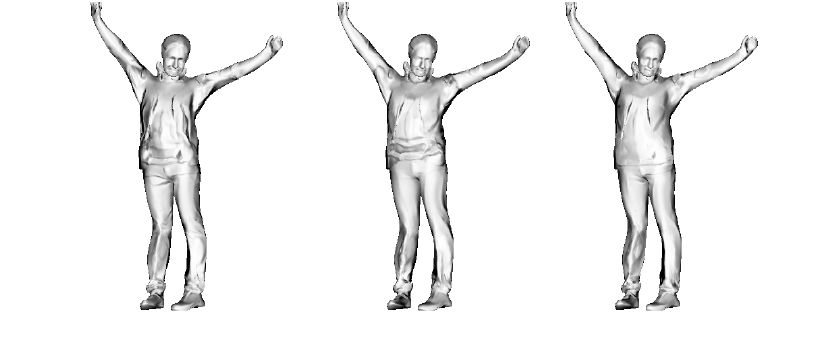}
    \vspace{-1.5em}
    \caption{\textbf{Disentangled Generation of Shape and Details.} We generate samples with the same coarse shape and different detail codes. Note that the details appear noticeably different from each other while all match the underlying coarse shape.}
    \label{fig:result_detail}
    \vspace{-1.5em}
\end{figure}

%% file: figures/result_nn.tex
\begin{figure}[t!]
    \centering
    \includegraphics[width=\linewidth,trim=15 15 20 0, clip]{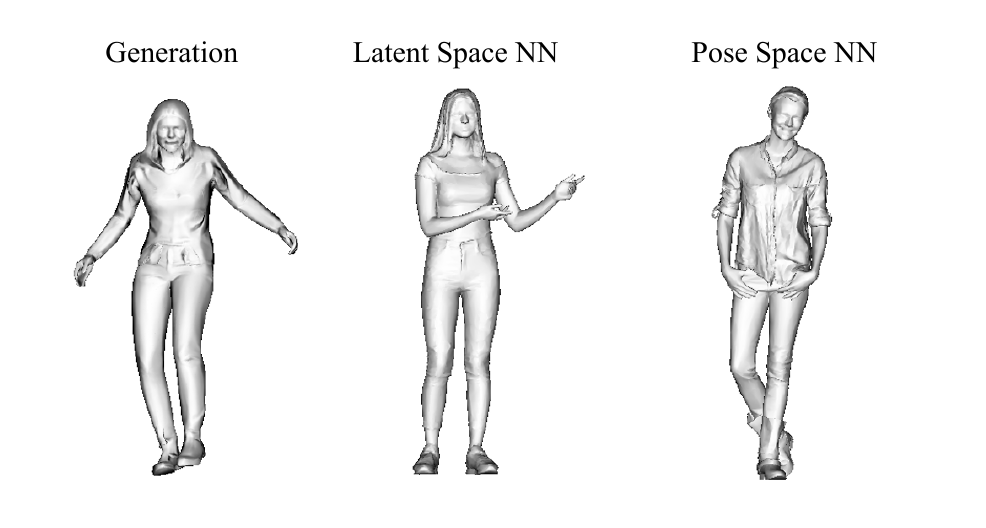}
    \caption{\textbf{Nearest Training Samples.} Note that the pose and clothing of the generated shape is noticeably different from that of the nearest training sample.}
    \label{fig:result_nn}
    \vspace{-0.4cm}
\end{figure}

%% file: tables/ablation.tex
\begin{table}
\centering
\resizebox{0.7\linewidth}{!}{ %
\begin{tabular}{@{}lccc@{}}
\toprule
Method & FID$\downarrow$ & User Preference$\uparrow$ \\ %
\midrule
Pose ONet & 43.80 & 8.11\% \\ %
Coarse Shape & 29.34 & 26.1\% \\ %
Detailed Normal (w/o Adv.) & 42.18 & 15.4\% \\ %
\textbf{Ours} & \textbf{11.54} & -- \\ %
\midrule
Ground-truth Scans & N/A & 78.7\% \\
\bottomrule
\end{tabular}
} %
\vspace{-0.2cm}
\caption{
\textbf{Ablation Study.} We report FID and user preference. The user preference score indicates how often participants of our perceptual study preferred a particular method over ours.
} %
\label{tab:ablations}
\vspace{-0.3cm}
\end{table}

%% file: figures/result_ablation.tex
\begin{figure}
    \centering
    \includegraphics[width=\linewidth,trim=0 45 300 20, clip]{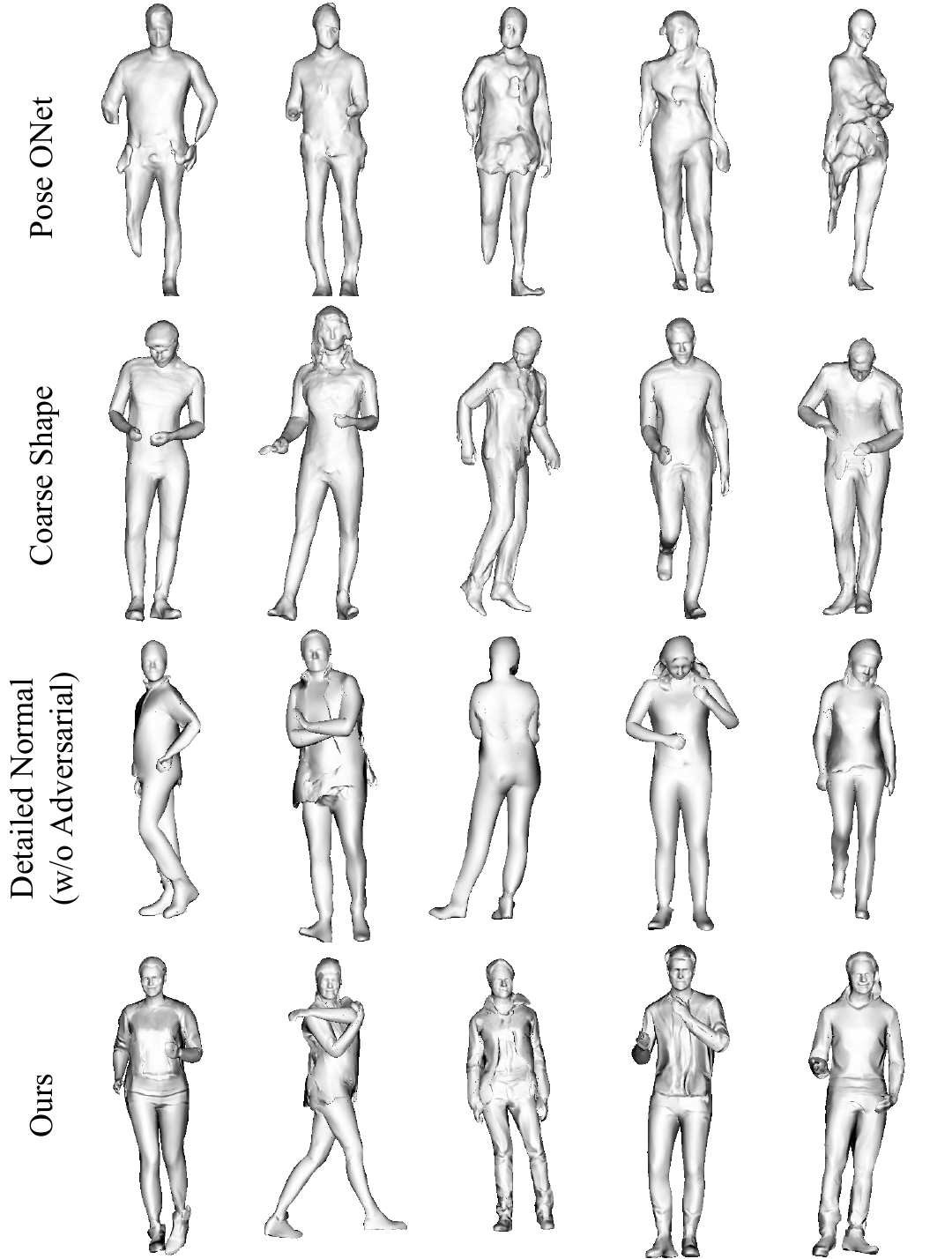}
    \vspace{-2em}
    \caption{\textbf{Generation Comparison.} We show random samples from ablative baselines and our method. %
    Without the adversarial loss, the generated shapes appear either bumpy (\emph{Coarse Shape}) or over-smooth (\emph{Detailed Normal w/o Adv.}).}
    \label{fig:result_ablation}
    \vspace{-0.5cm}
\end{figure}

%% file: sections/5_conclusion.tex
\section{Conclusion}
\vspace{-0.5em}
We propose \methodname, a generative model of 3D clothed humans that can produce a large variety of clothed people with detailed wrinkles and explicit pose control. Using implicit multi-subject forward skinning enables learning from only a few posed scans per subject. To model the stochastic details of garments, we exploit a 2D adversarial loss to update a 3D normal field.
We demonstrate that gDNA can be used in various applications such as animation and 3D fitting, outperforming state-of-the-art methods.

\update{\boldparagraph{Limitations} Learning loose clothing (\eg skirts) from deformed observations remains challenging due to the topology ambiguity and the large pose-dependent non-linear cloth deformation. Please refer to \suppmat for more discussions about limitations and societal impact.}

\input{tables/fitting}
\input{figures/result_repose_fitting}

%% file: tables/fitting.tex
\begin{table}
\centering
\resizebox{0.8\linewidth}{!}{ %
\begin{tabular}{@{}lccc@{}}
\toprule
Method & Pred-to-Scan $\downarrow$ & Scan-to-Pred $\downarrow$ \\
\midrule
SMPLicit~\cite{Corona2021CVPRSMPLicit} &N/A & 0.0240\\
NPMs~\cite{Palafox2021ICCV} & 0.0156 & 0.0215\\
\textbf{Ours} & \textbf{0.0134} & \textbf{0.0123}\\
\bottomrule
\end{tabular}
} %
\vspace{-0.5em}
\caption{
\textbf{Fitting Comparison.} We report the distance between the target scan and 3D shapes fit by SOTA methods and ours. Pred-to-Scan metric does not apply to multi-layer surfaces from SMPLicit.} %
\label{tab:fitting}
\vspace{-1em}
\end{table}

%% file: figures/result_repose_fitting.tex
\begin{figure}
    \centering
    \includegraphics[width=\linewidth,trim=20 145 0 15, clip]{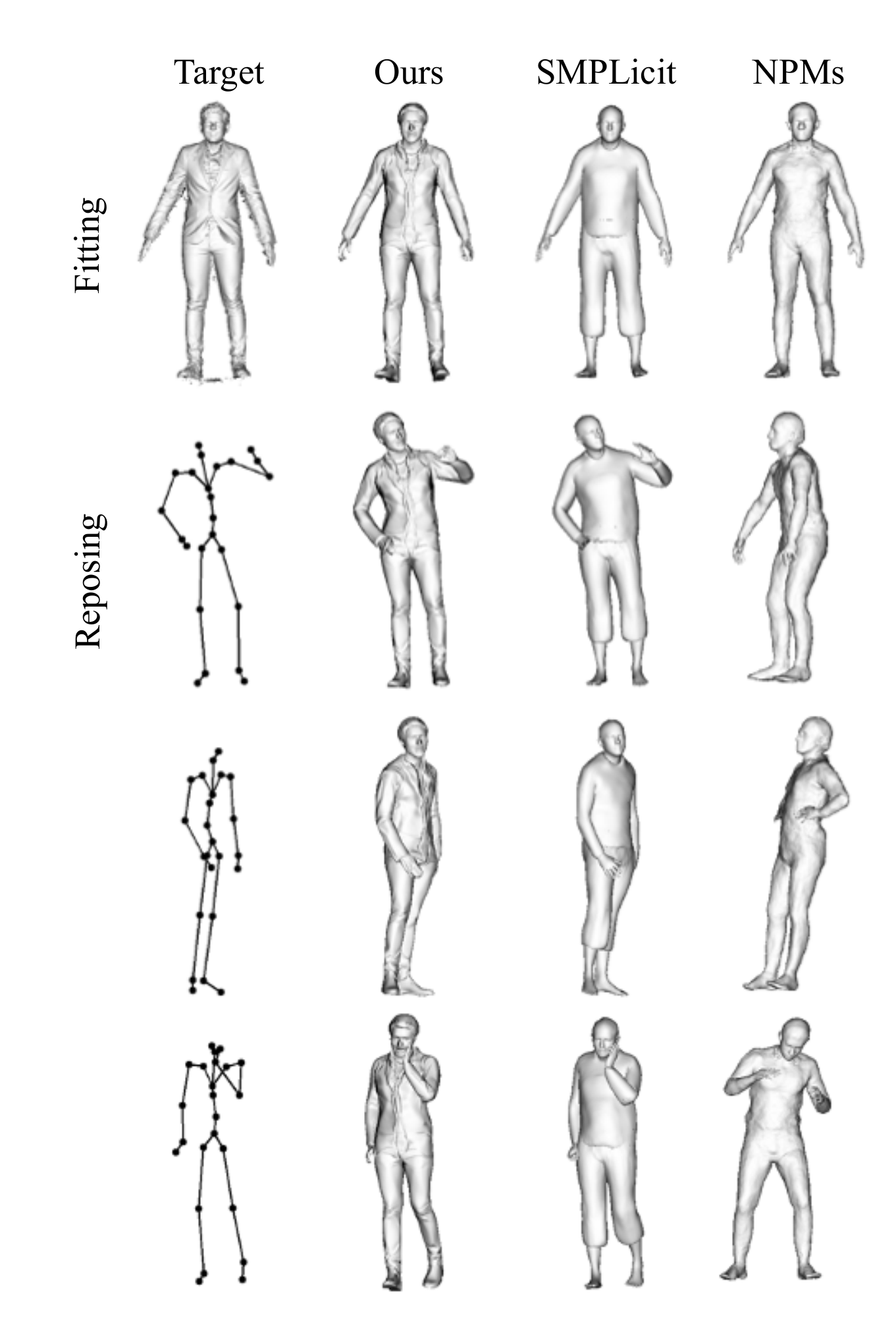}
    \vspace{-2em}
    \caption{\update{\textbf{Fitting and Reposing.} We compare model fitting results on the SIZER dataset with SMPLicit~\cite{Corona2021CVPRSMPLicit} and NPMs~\cite{Palafox2021ICCV}. We also show the fitted shapes reposed into target poses. As NPMs do not allow specifying target pose, random poses are shown.}}
    \label{fig:result_repose_fitting}
    \vspace{-1.7em}
\end{figure}

%% file: sections/6_acknowledgement.tex
\vspace{-0.2em}
\boldparagraph{Acknowledgements} Xu Chen was supported by the Max Planck ETH Center for Learning Systems. Andreas Geiger was supported by the DFG EXC number 2064/1 - project number 390727645. We thank Alex Zicong Fan, Marcel C. B\"uhler, Priyanka Patel, Qianli Ma, Sai Kumar Dwivedi, Thomas Langerak and Yuliang Xiu for their feedback, Garvita Tiwari for her suggestions about the SIZER dataset, and Tsvetelina Alexiadis for her help with the user study. 
\noindent\textbf{Disclosure:} 
MJB has received research gift funds from Adobe, Intel, Nvidia, Meta/Facebook, and Amazon.  MJB has financial interests in Amazon, Datagen Technologies, and Meshcapade GmbH.  MJB's research was performed solely at, and funded solely by, the Max Planck.